\documentclass{article}

\usepackage[preprint]{neurips_2019}

\usepackage[utf8]{inputenc} 
\usepackage[T1]{fontenc}    
\usepackage{hyperref}       
\usepackage{url}            
\usepackage{booktabs}       
\usepackage{amsfonts}       
\usepackage{nicefrac}       
\usepackage{microtype}      
\usepackage{adjustbox}      

\title{The State Of Knowledge Distillation For Classification Tasks}

\author {
    Fabian Ruffy\\
	frv217\\
	\texttt{fruffy@nyu.edu}\\
	\And
	Karanbir Chahal\\
	ksc487\\
	\texttt{karanchahal@nyu.edu}\\
}

\begin{document}

\maketitle

\begin{abstract}
    We survey various knowledge distillation (KD) strategies for simple classification tasks and implement a set of techniques that claim state-of-the-art accuracy. Our experiments using standardized model architectures, fixed compute budgets, and consistent training schedules indicate that many of these distillation results are hard to reproduce. This is especially apparent with methods using some form of feature distillation. Further examination reveals a lack of generalizability where these techniques may only succeed for specific architectures and training settings. We observe that appropriately tuned classical distillation in combination with a data augmentation training scheme gives an orthogonal improvement over other techniques. We validate this approach and open-source our code\footnote{\url{https://github.com/karanchahal/distiller}}.
\end{abstract}

\section{Introduction}
Modern deep neural networks are resource-intensive which limits their viability for bandwidth constrained and low power environments. Edge devices with their limited GPU size and battery power are hamstrung by the size of powerful deep learning models that run on most devices today. As a consequence, model compression, a discipline which seeks to decrease the size of a particular network while retaining accuracy has gained traction in recent years. 


A promising subclass of model compression is knowledge distillation which trains a smaller model (student) with low resource requirements on the logits of a larger model (teacher) on some particular task. These logits are posited as the "dark knowledge" of the teacher model and are meant to be an additional signal for the student to train on. These soft targets are easier to model for the student and have been shown to provide a boost to the final student accuracy.

In this work, we set out to survey the rich landscape of knowledge distillation and its gains in compression for image classification. We have three distinct aims in particular
\begin{enumerate}
    \item Implement the state-of-the-art in classification and validate top-performing methods.
    \item Understand how these approaches would work orthogonally.
    \item Assess the generalizability of each technique.
\end{enumerate}
We use the CIFAR10~\cite{cifar100} dataset as the benchmark for all our experiments and constrict all techniques to use a consistent optimizer, compute budget, and a single data augmentation scheme for all our experiments. 
Surprisingly, our experiments show that vanilla knowledge distillation performs the best after careful hyper parameter tuning. Most surprising of all, feature distillation techniques seem to perform worse than all other approaches. The biggest factor we encountered was, in fact, the difference in distillation performance when we switched to teachers of different architectures.

\section{Knowledge Distillation}
The most common knowledge distillation loss today was first introduced by Hinton et al.~\cite{hinton} in 2014. Knowledge Distillation, as proposed by Hinton et al., seeks to provide another pathway to gain knowledge about a task by training a model with a distillation loss in addition to the task loss. This distillation loss is generated by the help of a teacher network, which is "cumbersome", i.e., is large in size, but achieves high accuracy on a task. The objective of distillation is to increase the accuracy of a smaller network (the student) by aiding it's learning through this distillation loss. The precise KD loss is defined as:
\vspace{-2pt}
\begin{equation}
    \mathcal{L}_{KD} = (1-\alpha) CE(\hat{y}^S, y) + \alpha T^2 KL(\sigma((\hat{y}^T/T), \lambda((\hat{y}^S/T))
\end{equation}
where $CE$ refers to the conventional cross-entropy loss and $KL$ to the Kullback–Leibler divergence of the softmax $\sigma$, and the log-softmax $\lambda$. $T$ is the temperature, intended to smooth outputs from very large teacher models and $\alpha$ is a simple balancing weight.

\subsection{Extensions}
\label{sec:ext}
Over the years, many have modifications have taken the idea of knowledge distillation further. A particular line of research, initially developed by Romero et al., 2014~\cite{fitnets}, augments knowledge transfer by also considering the intermediate representation layers of the teacher as hints during the training process. Distilling the intermediate representation of task to a student is frequently referred to as feature distillation and is a burgeoning research area.

Most recently, bringing feature distillation to specific tasks and model types has received substantial attention. Various approaches have tried to distill the BERT model for language modelling and report a decrease in model size up to a factor of 7.5x with only a small decrease in accuracy~\cite{extreme_model_compression}. TinyBERT~\cite{tinybert} proposes a novel transformer distillation method which works by minimising the mean squared error between the various layers of a transformer network. They also propose a 2-step training pipeline which first trains a general BERT model and then along with a data augmentation technique fine-tuned this general model on downstream tasks. Ablation studies verify the effectiveness of this training approach and the importance of the distillation of multi-headed attention layers. Similar work by Sun et al., 2019~\cite{pkd_distillation}, referred to as "Patient Knowledge Distillation", distills BERT by minimising mean-squared error loss of each individual layer of the student and teacher.

In the field of image classification, Heo et al., 2019~\cite{overhaul_distillation} perform a comprehensive survey on the state of feature distillation. They surmise that their proposed method, which includes a feature transform with a margin ReLU, careful selection of the distillation feature position, and a partial L2 distance function surpasses their previous work (AB-Distillation~\cite{ab_distillation}) and results in state-of-the-art accuracy.

An approach that differs from previous distillation techniques is Relational Knowledge Distillation (RKD) by Park et al.,2019~\cite{rkd_distillation}. RKD distinguishes itself in that it focuses on the structural differences of teacher and student output instead of calculating the loss of the individual outputs. RKD introduces a distance loss and an angle loss that seeks to penalize the structural differences in relations. The team uses these losses in addition to a feature distillation loss and the Hinton loss with carefully tuned hyperparameters. 

Mirzadeh et al.,2019~\cite{takd_distillation} postulate that the student learns from a teacher that is far too advanced and hence fails to keep up. To remedy this, they first train an intermediate size "assistant" network by distilling the teacher. The student is then distilled down from this intermediate assistant network. It is shown that the student is able to achieve higher accuracy than baseline knowledge distillation using the transferred knowledge of the assistant.

\section{Baselines}
\label{sec:base}

We select a set of the high performing techniques described in Section~\ref{sec:ext} to estimate the status of knowledge distillation: Activation-Boundary Distillation (AB), Overhaul Distillation (OH), Relational Knowledge Distillation (RKD), and Teacher Assistant Distillation (TAKD). In the following section we detail how integrated each approach into our evaluation pipeline.

\subsubsection{Activation-Boundary Distillation (AB)}
Heo et al., 2019~\cite{ab_distillation} propose computing the distillation loss based on "activation boundaries" which are defined as the difference between positive and negative responses. This is opposed to comparing the absolute magnitude of the responses of two feature layers. We have selected AB-Distillation as the authors claim superior performance on CIFAR100 over the Hinton loss. 
Unfortunately, we experienced major difficulties in adopting AB-Distillation even thouigh we had access to the authors codebase. AB-Distillation has been tested on Wide-ResNets~\cite{wide_resnet} and it requires highly specific information from the neural network models (number of input channels per layer, access to concrete individual layers of the model, and each feature output). As a consequence, ResNet models had to be manually tweaked to fit with the proposed approaches. This entails choosing what hidden layers to distill from, matching feature map dimensions of the student/teacher due to the size difference and tuning hyperparameters for feature distillation. This specificity, which we encountered for many feature distillation research works, is a substantial downside of feature distillation.

The approach itself compares the feature values before ReLU is applied and uses a customized "hinge-like" loss on each layer. AB-Distillation consists of two phases. First, the student is initialized by aligning its features with the translated features of the teacher. After this the student is trained with standard knowledge distillation. We split the two phases into a 60:40 split as described in the paper. First we run the embedding initialization after which conventional knowledge distillation is trained. In preliminary experiments we observed marginal to no difference to KD, likely because the second distillation phase dominates. As a consequence, we decided to reimplement the follow-up to this work, Overhaul Distillation.

\subsubsection{Overhaul Distillation (OH)}
OH-Distillation operates similarly to AB-Distillation but improves by reconsidering the positions from where feature vectors are selected. OH picks a specific distillation position and computes a feature and activation boundary loss before a ReLU is applied. The hope is that by carefully picking the relevant ReLU layers and only considering activation boundaries, only the most necessary information is considered. We were able to integrate this technique to our evaluation pipeline by referring to the author's codebase\footnote{\url{https://github.com/clovaai/overhaul-distillation}} and by augmenting our ResNet models with the necessary adjustments to acquire this information.

\subsection{Relational Knowledge Distillation (RKD)}

RKD uses a variety of loss functions all of which seek to measure some relational properties about the vector. These ideas are represented in the form an angular distance loss, a distance-wise loss and a "dark-rank"- a measure typically used for metric learning. RKD also makes use of traditional KD, Attention-Transfer (AT)~\cite{at_distillation} and the FitNet~\cite{fitnets} loss. Using this loss ensemble, RKD claims top performance on Cifar100.
We re-implement the sections of RKD that do not require access to specific features of the model. This means we are not using AT and and FitNet loss. However, while this may hurt this overall performance, we still expect an improvement over conventional knowledge distillation. Unfortunately, the code for the CIFAR10 distillation used in the paper was not public at the time of this project\footnote{\url{https://github.com/lenscloth/RKD}}. We reimplement the approach based on descriptions of the paper and by reaching out to the authors and asking them for details.

\subsection{Teacher-Assistant Distillation (TAKD)}
Teacher-Assistant Knowledge Distillation~\cite{takd_distillation} follows a simple premise. Inspired by recent work~\cite{born_again}, which claims that self-distillation can actually improve accuracy of the base model, TAKD proposes multiple phases of distillation. Instead of training down from a single large teacher towards a student, TAKD uses a "teacher-assistant" (TA). The TA is typically a smaller model of the same architecture, which bridges the expressiveness gap between the teacher and student. The proposition is that the TA is able to translate classifications of the teacher which the student may not be able to express.

\section{New Techniques}
We develop three new variations of knowledge distillation which we detail below.

\subsubsection{Simple Feature Distillation (SFD)}
Patient Knowledge Distillation (PKD)~\cite{pkd_distillation} is a technique used for BERT model compression but is general enough that it to be translated into vision tasks. The fundamental idea is to simply minimise the mean-squared error between each individual layer of the student and teacher. Inspired by PKD, we implement a variant of this approach we call Simple Feature Distillation (SFD). SFD automatically retrieves all layers of the models and applies a max-pool operation across the teacher layer to bring it down to the size of the student layer. Max-pool compresses the feature layer while still retaining a strong signal. We experimented with deconvolutional, interpolation, and average pool layers and observe that it does not have significant difference.

Once feature layers are matched, the mean-squared error loss is minimised between the student and the teacher. The intuition behind this approach is to align the students activations with the compressed, translated representation of the teacher. We also experimented with minimising the KL divergence of the feature maps and observe no difference in final validation accuracy.   

\subsubsection{Ensemble Distillation (MKD)}
Based on the observations by Hinton and the TAKD paper we train a student under an ensemble of "generalist" teachers. This approach simply averages the Hinton KD-loss for each of the teacher and student outputs. The expectation is that the student will generalize better under a variety of diverse signals. We use multiple models of the same architecture to denote the ensemble of teachers.

\subsubsection{Unsupervised Distillation (UDA)}

Lastly, we also borrow techniques from unsupervised data augmentation learning~\cite{uda}. We built a training scheme where the data loader outputs an unaugmented image and an augmented image using the data augmentation policy RandAugment~\cite{randaugment} in the same mini batch. This policy works by randomly picking a transformation from a set of handpicked data augmentations known to work well with CIFAR10. We minimise the Hinton Knowledge Distillation loss between the teacher-student logits pairs for an augmented and an unaugmented image by summing them together. We call this loss the UDA-Cifar loss.

In another experiment, we concatenate the STL-10 and Cifar10 datasets. During training, on encountering the Cifar10 samples- the UDA-Cifar loss is used and on encountering the STL10 unsupervised samples- the UDA~\cite{uda} loss is used. The intuition for this approach is to use much more data and use the paradigm of self supervised training to squeeze out improvements. We make sure to have equal number of Cifar10 and STL samples in a single mini batch, as we observed the model performs worse if this hard example mining is not used. We term this method STL. This experiment was inspired by observing that a student distilled from a Cifar10 trained teacher using the STL dataset for distillation gives 83\% validation accuracy on Cifar10 inspite of never being trained on Cifar10.


\section{Experiments}
Our experiments consisted of three major phases. First, to understand the properties of knowledge distillation, we conduct a hyperparameter and model architecture analysis. Based on the results of the analysis we pick a teacher and parameter configuration and run all implemented techniques with the same configuration. We then pick the best performing techniques and average their performance over multiple extensive runs. 

\subsection{Setup}
\label{sec:setup}
The majority of our tests are performed on the CIFAR10~\cite{cifar100} dataset. We choose CIFAR because of a focus on image classification in the original work and due of its computational feasibility.\footnote{We also tried the same experiments using the larger, more complex CIFAR100 but the ultimate outcome remained the same. For the remainder of the paper, we will thus discuss only CIFAR10.} We can train CIFAR on smaller devices without requiring a multi-GPU setup. An important focus of our tests was to ensure fairness and even conditions. Correspondingly, all knowledge distillation experiments are run under the same conditions. 
All our tests are performed with the same optimizer. Stochastic Gradient Descent with Nesterov enabled, a momentum of 0.9 and an initial learning rate of 0.1 that is decreased by 0.1 at 33\% and 66\% of the total epochs. Weight decay is fixed to 0.0005. 
For training, we augment data using traditional CIFAR10 augmentation methods. We normalize each image, apply a random horizontal flip, and randomly crop the image to a size of 32x32 with a padding of 4. The validation set is only normalized.

The models we use are drawn from several different sources. Because the PyTorch sample architectures are designed for ImageNet~\cite{imagenet}, we primarily use the popular ResNet~\cite{resnet} architecture by the Github user Kuangliu\footnote{\url{https://github.com/kuangliu/pytorch-cifar}}, which has been optimized to achieve high accuracy on CIFAR10 and CIFAR100. These models quickly achieve >95\% accuracy on the validation set, which we deem more than sufficient. For knowledge distillation we use a heavily stripped version of this model, ResNet8. ResNet8 only contains three major blocks instead of four, and has substantially less parameters. Compared to our smallest 4-layer ResNet, ResNet10, which has 4,903,242 parameters and takes up 25.28MB of space, the ResNet8 only uses 89,322 parameters and takes up 2.88 MB of space in memory. We use the efficient ResNet8 for all of our knowledge distillation experiments. It achieves an approximate base accuracy of 89\%. we deem any technique that reliably achieves above 90\% a success.

\subsection{Parameter Tuning}
Before we conducted our CIFAR10 classification measurements we performed extensive hyperparameter search to understand tradeoffs in classical knowledge distillation. We asked ourselves two questions. 1) How do different combinations of $T$ and $\alpha$ affect distillation performance. 2) Does the architecture of the teacher matter for performance? 

For 1) we trained a ResNet8 with a Resnet26 teacher for 150 epochs for each combination of $\alpha = [0.1, 0.4, 0.5, 0.7, 1.0]$ and $T = [1, 5, 10, 15, 20]$. Table \ref{parameters} highlights selected results. The results show that knowledge distillation in our setup is relatively robust towards parameter selection with only marginal differences in performance. Even tests conducted with an $\alpha$ of 0.1 achieved significantly higher performance than normal training (\texttt{0.8814}).

Our second test evaluated the impact of model architecture on distillation performance. Again, we trained a ResNet8 for 150 epochs and selected various teacher architectures for comparison. Table \ref{teachers} highlights the results. 
Interestingly, model architectures have substantial impact on the final accuracy outcome. The highest accuracy is achieved by teachers (ResNet20, ResNet26) with the same structure, even though they may not have the highest classification accuracy. Wide-ResNets~\cite{wide_resnet} (WRN10-1 and WRN16-4) and 4-layer ResNets (ResNet18) are structurally different, which affects the student's ability to mimic the teacher outputs. We also ran an additional test for ResNet18 to assess if higher temperature values improve performance but we did not observe any noticeable effect. This may be because the parameter differences between the models do not exhibit sufficient magnitude.

Based on the results in 1) and 2) we ultimately decided to pick a ResNet26 teacher with 93.41 accuracy and train with an $\alpha$ of 0.5 and a $T$ of 5 for all subsequent tests.

\begin{table}
  \caption{Knowledge distillation results for different values of alpha.}
  \label{parameters}
  \centering
  \begin{tabular}{|c||c|c|c|c|c|c|c|c|c|}
    \hline
    Alpha & \multicolumn{3}{|c|}{0.1} &  \multicolumn{3}{|c|}{0.5} &  \multicolumn{3}{|c|}{1}  \\
    \hline
    Temperature &
    1  & 10 & 20 & 
    1  & 10 & 20 & 
    1  & 10 & 20  \\
    \hline
    Accuracy &
    89.72 & 89.90 & 90.01 & 
    90.33 & 90.18 & 89.91 &
    89.85 & 90.10 & \textbf{90.37} \\
    \hline
  \end{tabular}
\end{table}

\begin{table}
  \caption{Knowledge distillation results for different teacher models.}
  \label{teachers}
  \centering
  \begin{tabular}{|c||c|c|c|c|c|c|}
    \hline
    Teacher & ResNet8 & ResNet20 & ResNet26 &  ResNet18 & WRN10-1 & WRN16-4   \\
    \hline
    \hline

     Num Params & 89322 & 283754 & 380970 &  11173962 & 77850 & 2748890   \\
    \hline
     Layers & 31 & 67 & 85 & 62 & 34 & 56   \\
    \hline
    Teacher Accuracy & 89.59 & 93.02 & 93.41 &  95.26 & 88.06 & \textbf{95.42}   \\
    \hline
    Student Accuracy & 89.48 & 90.16 & \textbf{90.40} &  89.50 & 88.48 & 89.58   \\
    \hline
  \end{tabular}
\end{table}

\subsection{Cifar10 Classification Experiments}
\subsubsection{Preliminary Comparisons}
We run each of the implemented techniques for 200 epochs using the configuration described in Section \ref{sec:setup}. For each run, we collect the highest validation accuracy achieved. Table \ref{experiments} shows the results. Unfortunately, all feature distillation techniques underperform severely and are incapable of even beating normal training. Only RKD outperforms the baseline, but this is likely due to the added knowledge distillation loss, which is missing in SFD and OH. 

\begin{table}
  \caption{Initial test run for 200 epochs. Results sorted in descending order.}
  \label{experiments}
  \begin{adjustbox}{center}
  \begin{tabular}{|c||c|c|c|c|c|c|c|c|c|c|c|}
    \hline
    Method & Teacher & TAKD & UDA & MKD & KD & RKD & NOKD & SFD & OH &STL  \\
    \hline
    \hline
    Accuracy & 93.41 & 90.57 & 90.52 & 90.34 & 90.33 & 90.22 & 89.42 & 87.47 & 87.09 & 90.8  \\
    \hline
  \end{tabular}
  \end{adjustbox}
\end{table}

\subsection{In-depth Analysis}
We investigate all techniques which outperform knowledge distillation more thoroughly. We removed MKD because it failed to achieve consistently higher performance than KD. We reran the experiment tbree times with 350 iterations and average the results of each technique. Table~\ref{second_run} shows the final results. UDA Distillation clearly gains an edge over TAKD and normal KD, meaning that the unsupervised data augmentation loss can provide substantial benefits. To verify that this performance boost was not simply because of regular data augmentation, we also ran a UDA test with the Hinton loss. While it achieved better performance that normal training, the improvement over KD in final performance was still significant. We speculate that simple knowledge distillation paired with sophisticated augmentation and hyper-parameter tuning can match top-accuracy feature distillation techniques. Unfortunately, because we were unable to reproduce "working" feature distillation methods in time, we were unable to confirm this hypothesis. 


\subsubsection{Why does Feature Distillation Perform so Poorly?}
We thoroughly investigated our code and reached out to authors for clarification, but could not identify the exact reason for the subpar performance. A recent work by Tian et al., 2019~\cite{crd_distillation}, which implemented 10 different distillation approaches, echoes our underwhelming experience. The authors found that all of them fail to outperform the knowledge distillation baseline. These and our results indicate that many feature distillation techniques may not generalize and achieve underwhelming results when translated into a slightly different context. We hypothesize that this is due to each method apart from KD for a particular model architecture, size, and training scheme. Our baseline models are comparably small and differ slightly in design from the models in the original source code. Another potential reason could be a subtle bug in the implementation of the Kullback–Leibler divergence in PyTorch\footnote{\url{https://github.com/pytorch/pytorch/issues/6622}}, which may have affected measurements. We uncovered this bug while implementing knowledge distillation and adopting the proposed fix lead to ~1\% KD performance increase across all KD experiments. Older feature distillation experiments may have compared against the faulty implemented divergence loss. 

Based on these observations, we are investigating the structural properties of the models and how feature distillation performance may be affected.

\subsection{Failures}
Here we detail some approaches that did not work.

\begin{table}
  \caption{Second test run for 350 epochs.}
  \label{second_run}
  \begin{adjustbox}{center}
  \begin{tabular}{|c||c|c|c|c|c|c|}
    \hline
    Method & Teacher & UDA & TAKD & KD & UDA-NOKD & NOKD  \\
    \hline
    \hline
    Accuracy & 93.41 & 91.22 & 90.97 & 90.77 & 90.34 & 89.34  \\
    \hline
  \end{tabular}
  \end{adjustbox}
\end{table}

\subsubsection{Dual Training}

During dual training, the teacher is trained until it achieves an accuracy greater than 2\% than the student. At this point, the teacher is frozen and the student is trained using the KD loss until it reaches or surpasses the teacher's accuracy. We observe that the student does not get a better final validation accuracy than just performing regular KD on a pre-trained teacher.

\subsubsection{Unsupervised Knowledge Distillation}

We trained the student the STL-10 dataset by simply minimising the KL divergence between the student and teacher logits on the large dataset. We observe that the student does learn and gains a validation accuracy of 83\% on Cifar10 inspite of never being trained on it.
Exploring the limits of knowledge distillation with a large amount of unsupervised data could be an interesting direction for future research.
\subsection{Experiments in Feature Distillation}

We tried a lot of different loss functions such as a cosine embedding loss, mean square error loss and KL divergence loss on the feature distillation but none of them improved accuracy. We also tried different combinations of matching the feature layers of the student and the teacher such as take only N last layers, take only N first layers, skip every N layers along with extensive hyperparameter tuning. All of our experiments in feature distillation ended in disappointment. One approach that seems intuitive is to distill fine grained pruned models into a small student. As the feature maps there would have a majority of zeros, they would theoretically be easier for the student to model.

\section{Conclusion}
We observe that a resnet18 achieving an accuracy of 95\% is able to distill a resnet8 student to an accuracy of 88.5\%. However, switching to a resnet26 teacher who's final validation accuracy was 93\% allowed for much more effective distillation. It resulted in the same resnet8 student achieving an accuracy of 90.5\% which results in close to a 2\% increase in accuracy.

We also observe that feature distillation is a hard problem and all our experiments with state of the art approaches performed worse than baseline performance of the student. There is an idea that the student might be too small to model the problem effectively. Our model might simply have hit it's parametric modelling limit and is unable to get a better accuracy.

We plan to explore a combination of pruning and feature distillation and understand how to identify the parametric modelling limits of a neural network in future work.

{\small
	\bibliographystyle{acm}
	\bibliography{chahal_ruffy_cv_project_2019}

\begin{thebibliography}{10}

\bibitem{randaugment}
{\sc Cubuk, E.~D., Zoph, B., Shlens, J., and Le, Q.~V.}
\newblock Randaugment: Practical data augmentation with no separate search.
\newblock {\em arXiv preprint arXiv:1909.13719\/} (2019).

\bibitem{imagenet}
{\sc Deng, J., Dong, W., Socher, R., Li, L.-J., Li, K., and Fei-Fei, L.}
\newblock {ImageNet: A Large-Scale Hierarchical Image Database}.
\newblock In {\em CVPR09\/} (2009).

\bibitem{born_again}
{\sc Furlanello, T., Lipton, Z.~C., Tschannen, M., Itti, L., and Anandkumar,
  A.}
\newblock Born again neural networks.
\newblock {\em arXiv preprint arXiv:1805.04770\/} (2018).

\bibitem{resnet}
{\sc He, K., Zhang, X., Ren, S., and Sun, J.}
\newblock Deep residual learning for image recognition.
\newblock In {\em Proceedings of the IEEE conference on computer vision and
  pattern recognition\/} (2016), pp.~770--778.

\bibitem{overhaul_distillation}
{\sc Heo, B., Kim, J., Yun, S., Park, H., Kwak, N., and Choi, J.~Y.}
\newblock A comprehensive overhaul of feature distillation.
\newblock {\em arXiv preprint arXiv:1904.01866\/} (2019).

\bibitem{ab_distillation}
{\sc Heo, B., Lee, M., Yun, S., and Choi, J.~Y.}
\newblock Knowledge transfer via distillation of activation boundaries formed
  by hidden neurons.
\newblock In {\em Proceedings of the AAAI Conference on Artificial
  Intelligence\/} (2019), vol.~33, pp.~3779--3787.

\bibitem{hinton}
{\sc Hinton, G., Vinyals, O., and Dean, J.}
\newblock Distilling the knowledge in a neural network.
\newblock {\em arXiv preprint arXiv:1503.02531\/} (2015).

\bibitem{tinybert}
{\sc Jiao, X., Yin, Y., Shang, L., Jiang, X., Chen, X., Li, L., Wang, F., and
  Liu, Q.}
\newblock Tinybert: Distilling bert for natural language understanding.
\newblock {\em arXiv preprint arXiv:1909.10351\/} (2019).

\bibitem{cifar100}
{\sc Krizhevsky, A., Hinton, G., et~al.}
\newblock Learning multiple layers of features from tiny images.
\newblock Tech. rep., Citeseer, 2009.

\bibitem{takd_distillation}
{\sc Mirzadeh, S.-I., Farajtabar, M., Li, A., and Ghasemzadeh, H.}
\newblock Improved knowledge distillation via teacher assistant: Bridging the
  gap between student and teacher.
\newblock {\em arXiv preprint arXiv:1902.03393\/} (2019).

\bibitem{rkd_distillation}
{\sc Park, W., Kim, D., Lu, Y., and Cho, M.}
\newblock Relational knowledge distillation.
\newblock In {\em Proceedings of the IEEE Conference on Computer Vision and
  Pattern Recognition\/} (2019), pp.~3967--3976.

\bibitem{fitnets}
{\sc Romero, A., Ballas, N., Kahou, S.~E., Chassang, A., Gatta, C., and Bengio,
  Y.}
\newblock Fitnets: Hints for thin deep nets.
\newblock {\em arXiv preprint arXiv:1412.6550\/} (2014).

\bibitem{pkd_distillation}
{\sc Sun, S., Cheng, Y., Gan, Z., and Liu, J.}
\newblock Patient knowledge distillation for bert model compression.
\newblock {\em arXiv preprint arXiv:1908.09355\/} (2019).

\bibitem{crd_distillation}
{\sc Tian, Y., Krishnan, D., and Isola, P.}
\newblock Contrastive representation distillation.
\newblock {\em arXiv preprint arXiv:1910.10699\/} (2019).

\bibitem{uda}
{\sc Xie, Q., Dai, Z., Hovy, E., Luong, M.-T., and Le, Q.~V.}
\newblock Unsupervised data augmentation.
\newblock {\em arXiv preprint arXiv:1904.12848\/} (2019).

\bibitem{at_distillation}
{\sc Zagoruyko, S., and Komodakis, N.}
\newblock Paying more attention to attention: Improving the performance of
  convolutional neural networks via attention transfer.
\newblock {\em arXiv preprint arXiv:1612.03928\/} (2016).

\bibitem{wide_resnet}
{\sc Zagoruyko, S., and Komodakis, N.}
\newblock Wide residual networks.
\newblock {\em arXiv preprint arXiv:1605.07146\/} (2016).

\bibitem{extreme_model_compression}
{\sc Zhao, S., Gupta, R., Song, Y., and Zhou, D.}
\newblock Extreme language model compression with optimal subwords and shared
  projections.
\newblock {\em arXiv preprint arXiv:1909.11687\/} (2019).

\end{thebibliography}
}

\end{document}